\documentclass[sigconf]{acmart}

\usepackage{algorithm}% http://ctan.org/pkg/algorithms
\usepackage{natbib}
\usepackage{algorithmic}
\usepackage{amsmath}
\AtBeginDocument{%
  \providecommand\BibTeX{{%
    \normalfont B\kern-0.5em{\scshape i\kern-0.25em b}\kern-0.8em\TeX}}}

\setcopyright{none}
\copyrightyear{MMXIX}
\definecolor{AH}{RGB}{0,0,0}
\newcommand{\AH}[1]{\textcolor{AH}{#1}}
\begin{document}

\title{An adaptive simulated annealing EM algorithm for inference on non-homogeneous hidden Markov models}

\author{Aliaksandr Hubin}
\email{aliaksandr.hubin@nr.no}
\orcid{1234-5678-9012}
%\email{webmaster@marysville-ohio.com}
\affiliation{%
  \institution{Norwegian Computing Center}
  \streetaddress{P.O. Box 114 Blindern}
  \city{Oslo}
  \state{Norway}
  \postcode{NO-0314}
}

%\author{blinded}
%\email{blinded}
%\email{webmaster@marysville-ohio.com}
%\affiliation{%
%  \institution{blinded}
%  \streetaddress{blinded}
%  \city{blinded}
%  \state{blinded}
%  \postcode{blinded}
%}

% \affiliation{%

\renewcommand{\shortauthors}{Hubin et.al.}

\begin{abstract}
  Non-homogeneous hidden Markov models (NHHMM) are a subclass of dependent mixture models used for semi-supervised learning, where both transition probabilities between the latent states and mean parameter of the probability distribution of the responses (for a given state) depend on the set of $p$ covariates. A priori we do not know which (and how) covariates influence the transition probabilities and the mean parameters. This induces a complex combinatorial optimization problem for model selection with $4^p$ potential configurations. To address the problem, in this article we propose an adaptive  (A) simulated annealing (SA) expectation maximization  (EM) algorithm (ASA-EM) for joint optimization of models and their parameters with respect to a criterion of interest.
\end{abstract}

%\ Show the XML Only




\maketitle

\section{Introduction}
Hidden Markov models (HMM) represent a tool used for pattern recognition, inference and predictions in both social and natural sciences, in different areas and for different purposes.  For example, in economics and econometrics, they are used for modelling regime-switching processes \cite{erlwein2010hmm} and various jump processes \cite{elliott2012hmm}. In psychology, they are used for learning process modelling \cite{wickens1982models}.  Other applications include speech recognition \cite{gales1998maximum}, biology and genetics \cite{krogh1994hidden}. \AH{Non-homogeneous hidden Markov models (NHHMMs) are the extension of standard HMMs that allow to model both transition probabilities between the latent states and mean parameters of the probability distribution of the responses with respect to a set of covariates \citep{hughes1999non}. NHHMMs typically have the same applications as standard HMMs \citep{maruotti2012mixed}, but allow additional flexibility. In this article, we address the model (variable) selection problem in NHHMMs. Model selection problem in terms of the choice of observations influenced by the transitions of a hidden Markov chain is well studied for simple HMMs and hidden semi-Markov models (HSMMs). Existing methods include variational Bayes (VB) \citep{zhu2012simultaneous}, Markov chain Monte Carlo (MCMC) \citep{boys2001comparison} and adaptations of expectation maximization (EM) \citep{adams2016feature}, and are summarized in \citet{adams2019survey}. At the same time, the inverse problem of selecting exogenous variables that, conditionally on the hidden state, influence the distributions of the observations as well as the problem of selecting exogenous variables influencing transition probabilities are much less studied. In \citet{paroli2007bayesian}, a comparison of different strategies of variable selection in NHHMMs is given, however, there only Gaussian observations are considered, the problem is limited to only inference (with no guidance to predictions), and only variable selection for transition probabilities is allowed. The extended MCMC based approach by the same authors is considered in \citet{paroli2008bayesian}, where both variable selection for transitions and observations as well as the problem of selecting the number of hidden states are studied, however, the approach is still limited in the sense that only Gaussian observations are addressed and no guidance for model-based predictions is given. A more recent prediction driven approach based on reversible jump MCMC \citep{green2009reversible} is suggested in \citet{meligkotsidou2011forecasting}, however, there, also, only Gaussian observations are considered with variable selection only on the level of transition probabilities and no guidance for pattern recognition driven applications. All of the approaches mentioned above are certainly sound mathematically and can be extended to broader problems with a moderate effort, however, they (as of today) remain slightly impractical for several reasons: 1. They are computationally heavy; 2. They are not designed for parallel computing; 3. No implementations or software are publically available for the scientific community; 4. Only Gaussian observations are considered. This results in that a) the approaches are only suited for small problems (6 covariates at most were considered in application of \citep{paroli2007bayesian, paroli2008bayesian, meligkotsidou2011forecasting}), b) even more recent works cannot easily adapt them and still rely on manual variable selection \citep{holsclaw2017bayesian}, c) it is impossible to benchmark against them on new data sets without reimplementing the approaches from scratch.}

\AH{The main contribution of this paper is a pragmatic, yet rigorous, solution for variable and model selection problems in NHHMMs. We consider any distribution of the observations from the exponential family and address both pattern recognition and prediction driven examples. Additionally, we enable practical parallel computing (allowing to perform variable selection for much larger sets of covariates) and provide an R package on GitHub~\footnote{\url{https://github.com/aliaksah/depmixS4pp}} for common use.} In particular, we suggest an adaptive irreducible simulated annealing \AH{expectation maximization algorithm (ASA-EM) and its parallel version} for joint optimization of models and their parameters with respect to a criterion of interest, which can be chosen among Akaike information criterion (AIC), Bayesian information criterion (BIC), Deviance information criterion (DIC), focused information criterion (FIC), \AH{marginal log-likelihood (MLIK), marginal posterior model probability (PMP) and others \cite{claeskens2008model}.} Alternatively, a time-aware cross-validation can be used. \AH{All of the mentioned model selection strategies induce regularization on the model complexity and hence prevent the model from overfitting. The suggested ASA-EM algorithm has an asymptotic guarantee to find the optimal model and values of its parameters, however, as we will show in the experiments section, it works well even for the limited time computations. To illustrate its performance, we address an example for financial stock prices data from \emph{S\&P 500} (30 covariates), and an example dealing with epigenetic observations from Arabidopsis  plant (17 covariates). Both of the examples address significantly larger sets of covariates than reported in all of the previous studies \citep{paroli2007bayesian, paroli2008bayesian, meligkotsidou2011forecasting}.}

\section{Mathematical model}

We assume a parametric probability distribution $\mathfrak{f}$ of the responses $Y_i, i \in \{1,...,n\}$, which come from the exponential family \cite{claeskens2008model}. We define separate mean ($\mu_{s_i}$) and dispersion ($\phi_{s_i}$) parameters, conditional on the latent states $s_i \in \{1,...,S\}$, which are assumed to have a Markovian dependence structure along indices $\{1,...,n\}$ allowing to incorporate a simple \textit{temporal} dependence between the observations:
\begin{align}
  &Y_i|\mu_{s_i},\phi_{s_i},s_i\sim   \mathfrak{f}(y|\mu_{s_i},\phi_{s_i}),
  \quad \mu_{{s_i}} = g^{-1}\left(\eta_{s_i}\right), \label{themodelbeg}\\
   &\eta_{s_i} =  \beta_0^{(s_i)} + \sum_{j=1}^{p} \gamma_j\beta^{(s_i)}_{j}x_{ij},\\
   &p(s_i|s_{i-1})=\frac{e^{\omega_0^{(s_i,s_{i-1})} + \sum_{j=1}^{p} \delta_j\omega^{(s_i,s_{i-1})}_{j}x_{ij}}}{1+e^{\omega_0^{(s_i,s_{i-1})} + \sum_{j=1}^{p} \delta_j\omega^{(s_i,s_{i-1})}_{j}x_{ij}}}. \label{themodeleqend}
\end{align}
Here, $g(\cdot)$ is the link function, $\beta^{(s_i)}_j \in \mathbb{R}, j \in \{0,...,p\}$ are regression coefficients of the covariates of the model showing for a given state $s_i$ whether and how the corresponding covariate influences the mean parameter $\mu_{s_i}$ of the distribution of the responses, $\gamma_i \in \{0,1\}, i \in \{1,...,M\}$ are latent indicators, defining if (for any state) covariate $i$ is included into the linear predictor of the model ($\gamma_i = 1$) or not ($\gamma_i = 0$).  $\omega^{(s_i),(s_{i-1})}_j \in \mathbb{R}, j \in \{0,...,p\}$ are regression coefficients of the covariates of the model showing for a given state $s_{i-1}$ whether and in which way the corresponding covariate influences the transition probabilities to state $s_{i}$. $\delta_i \in \{0,1\}, i \in \{1,...,p\}$ are latent indicators, defining if (for any state) covariate $i$ is included into the model for transition probabilities between the latent states ($\delta_i = 1$) or not ($\delta_i = 0$). Here both the mean parameters of the responses and the transition probabilities of the Markov chain depend upon up to $p$ covariates and the combinatorics of the choice of the covariates (model configurations) is incorporated via the latent binary indicators, which switch the covariates on and off separately for the latent Markov chain of the states (through $\delta_i$) and the distribution of the responses (through $\gamma_i  $). At the same time, to avoid super-exponential explosion of the number of candidate models, we assume that the same variables are influencing the observations for every hidden state and the same pattern of variables is influencing transition probabilities between all pairs of hidden states. The following section will present the algorithm for fitting the model \eqref{themodelbeg}-\eqref{themodeleqend}.

\section{Adaptive simulated annealing EM algorithm}
In order to deal with the presence of latent states and latent binary indicators for covariate selection as well as multimodality in the joint space of model configurations  $\boldsymbol m = \{\boldsymbol\delta,\boldsymbol\gamma\}\in \mathcal{M}$ and their parameters $\boldsymbol \theta_m = \{\boldsymbol\beta,\boldsymbol\phi|\boldsymbol m\}\in \boldsymbol{\Theta}_m$ we will combine the adaptive simulated annealing (SA) algorithm (for model space exploration) and the standard expectation maximisation (EM) algorithm (for parameter space exploration) into the novel ASA-EM algorithm, described in Algorithm~\ref{a1}. There we consider the objective criterion $r(\boldsymbol m,\boldsymbol\theta_m|\boldsymbol Y, \boldsymbol x)$ to be minimized across $\boldsymbol m \in \mathcal{M}$ and $\boldsymbol \theta_m \in \boldsymbol{\Theta}_m$. The objective $r(\boldsymbol m,\boldsymbol\theta_m|\boldsymbol Y, \boldsymbol x)$ can be chosen as AIC, BIC, DIC, FIC, MLIK or PMP  \cite{claeskens2008model}, with no loss of generality. Acceptance ratio of ASA-EM Algorithm~\ref{a1} at temperature $\tau$ is defined as
\begin{equation}
a_\tau(r_1,r_2) = \min\left\{1,\exp{\frac{\left(r_1-r_2\right)}{\tau}}\right\}.\label{algACC}
\end{equation}
Here $r_1=r(\boldsymbol m_1,\boldsymbol\theta_{m_1}|\boldsymbol Y, \boldsymbol x)$ and $r_2=r(\boldsymbol m_2,\boldsymbol\theta_{m_2}|\boldsymbol Y, \boldsymbol x)$.

\subsection{The EM algorithm}
In this subsection, we describe the EM part of the ASA-EM algorithm in more detail.

Conditional on having a fixed model $m$ from the model space $\mathcal{M}$, the standard expectation maximization (EM) algorithm is used to make inference on the parameters $\boldsymbol\theta_m$ of the model $m$ \citep{krogh1994hidden}. The inference algorithm here works as follows: One first introduces the likelihood of the model parameters $p(\boldsymbol{Y}|\boldsymbol{x},\boldsymbol\theta_m)$. Here, however, the parameters of the model are not defined without the sequence of latent states $\boldsymbol{s}$ for all of the observations. We hence add them and then integrate out using the law of total probability as: \begin{displaymath}p(\boldsymbol{Y}|\boldsymbol{x},\boldsymbol\theta_m) = \sum_{\boldsymbol{s}\in \Omega}p(\boldsymbol{Y},\boldsymbol{s}|\boldsymbol{x},\boldsymbol\theta_m).\end{displaymath} At this point, one still does not know explicitly $p(\boldsymbol{Y},\boldsymbol{s}|\boldsymbol{x},\boldsymbol\theta_m)$. To resolve this, one can multiply and divide the expression by some auxiliary $q(\boldsymbol{s}|\boldsymbol\theta_m)$: \begin{displaymath}p(\boldsymbol{Y}|\boldsymbol{x},\boldsymbol\theta_m) = \sum_{\boldsymbol{s}\in \Omega}q(\boldsymbol{s}|\boldsymbol\theta_m)\frac{p(\boldsymbol{Y},\boldsymbol{s}|\boldsymbol{x},\boldsymbol\theta_m)}{q(\boldsymbol{s}|\boldsymbol\theta_m)}.\end{displaymath} 
In the E-step of the EM algorithm one obtains the likelihood explicitly by taking expectation with respect to $q(\boldsymbol{s}|\boldsymbol\theta_m^*)$. This means getting: 
\begin{equation}
p(\boldsymbol{Y}|\boldsymbol{x},\boldsymbol\theta_m) =  \mathbb{E}_{q(\boldsymbol{s}|\boldsymbol\theta_m^*)}\left(\frac{p(\boldsymbol{Y},\boldsymbol{s}|\boldsymbol{x},\boldsymbol\theta_m)}{q(\boldsymbol{s}|\boldsymbol\theta_m^*)}\right).\label{expEM}
\end{equation} 
Here, $\boldsymbol\theta_m^*$ are the values of parameters obtained at a previous M-step. Further the M-step consists of maximizing the expectation \eqref{expEM} with respect to $\boldsymbol\theta_m$ to obtain the new $\boldsymbol\theta_m^*$, which will be used in the new EM iteration. Typically, $q(\boldsymbol{s}|\boldsymbol\theta_m)$ utilizes the Markovian properties of the latent sequence of states to make the algorithm computationally more efficient. Also note that in the E-step, analytical expectation can be exchanged with importance sampling, if the former is difficult to obtain.

\subsection{SA model exploration algorithm}
Having the methodology for estimating $\boldsymbol{\theta}_m$ within the model, we can move on to the model selection algorithm, which is an adaptive simulated annealing procedure. We will here first describe the standard SA with fixed hyperparameters.

The standard simulated annealing algorithm generates a sequence of \textit{solutions} $\{z_1,...,z_k\}$ minimizing some objective $\mathsf{G}(z)$~\footnote{In our case the solutions are different models and their parameters, i.e. $z_i = \{\boldsymbol{m}_i,\boldsymbol{\theta}_{m_i}\}$, but we will in this section address $z_i$, because the focus for the reader here is understanding the standard simulated annealing procedure. Additionally for simplicity we consider a general objective $\mathsf{G}(z)$ to be minimized, whereas the focus of ASA-EM will be a specific $\mathsf{G}(z):=r(\boldsymbol m,\boldsymbol\theta_m|\boldsymbol Y, \boldsymbol x)$} as described in Algorithm~\ref{sa1}.
\begin{algorithm}[H]
\tiny
   \caption{\small Standard simulated annealing optimization}
   \label{sa1}
\begin{algorithmic}
	\STATE  $z\gets z_0$;
	\STATE  $z_b\gets z_0$;
	\FOR{$\tau$ {\bfseries in} $\{\tau_{max},...,\tau_{min}\}$} %\Comment{for all temperatures in the cooling schedule}
	 \FOR{$k$ {\bfseries in} $\{1,...,K\}$}
	\STATE  $z_c\gets \mathbb{N}(z)$; %\Comment{pick a random neighbor of the current solution}
	\IF{$a_\tau(\mathsf{G}(z),\mathsf{G}(z_c))\geq u \sim Unif[0;1]$}
		\STATE  $z \gets z_c$; %\Comment{accept the move with some probability}
		\IF{${G}(z_c)<{G}(z_b)$}
	\STATE  $z_b\gets z_c$; %\Comment{update the best found solution}
	\ENDIF
	\ENDIF
	\ENDFOR 
	\ENDFOR 
   \STATE \textbf{return} $z, z_b$%\Comment{return the final solution}
\end{algorithmic}
\end{algorithm}
Here the cooling schedule is assumed to be exponential, i.e. \begin{displaymath} \tau_{i+1}=\tau_i\exp{(-\kappa_t)}.\end{displaymath} Acceptance probabilities for the simulated annealing kernel are of form~\eqref{algACC} but with $r_1=\mathsf{G}(z_1)$ and $r_2=\mathsf{G}(z_2)$.
The initial point $z_0$ is drawn randomly.
The limiting distributions of SA for each temperature $\tau$ are found (due to the Markovian property of the procedure) as follows:
\begin{equation}
p(z_k=z|\tau)\propto\exp{\left(-\frac{\mathsf{G}(z)}{\tau}\right)}.
\end{equation}
Note that if $\mathsf{G}(z)$ is the likelihood multiplied by the prior scaled with $\tau$, i.e. for $\mathsf{G}(z) = -\tau\log p(\boldsymbol{Y}|\boldsymbol{x},z)p(z)$ the SA corresponds to the Metropolis-Hastings MCMC algorithm \citep{geyer1995annealing} and hence gives samples from the posterior distribution of $z$. This can be easily utilized in our approach if one is \AH{interested} in the whole \textit{posterior distribution of models and parameters} rather than one \textit{best} solution.

\subsection{Adaptive SA hyperparameter tuning}
Finally, the adaptive part of the ASA-EM acronym means that the algorithm is gradually learning its own hyperparameters over some time before switching to a stationary regime after $R_a$ epochs. The tuning process is described in this subsection.

$\blacklozenge$ For a random variable of the number of iterations of EM algorithm $l$ we set $l \sim \text{Poisson}(\lambda)$ with the parameter prior of the form $\lambda\sim\text{Gamma}(a_\lambda,b_\lambda)$, which is conjugate to the distribution of $l$. Then throughout training, we draw proposals $l$ from the following posterior distribution: 
\begin{equation}
 p(l|D_l)\overset{d}{=}\text{NB}\left(a_\lambda +\sum_{x\in D_l}{x} ,\frac{1}{1 + b_\lambda + ||D_l||}\right),   
\end{equation}
which we call \textit{adaptive posterior distribution}. Here $||\Vec{v}||$ is a $L_0$ norm representing the length of the vector $\Vec{v}$ and $D_l$ is the synthetic data on the (what we call) \textit{successful} iterations, that is if a new global optima in terms of $r(\boldsymbol m,\boldsymbol\theta_m|\boldsymbol Y, \boldsymbol x)$ or the move is accepted by SA algorithm at temperature $\tau<1$, we append the current $l$ to $D_l$. 

$\blacklozenge$ The number of iterations $K$ of SA per temperature $\tau$ is also assumed Poisson distributed $K \sim \text{Poisson}(\mu)$ with a conjugate prior $\mu\sim\text{Gamma}(a_\mu,b_\mu)$. During training we draw proposals $K$ from the adaptive posterior distribution:
\begin{equation}
p(K|D_k)\overset{d}{=}\text{NB}\left(a_\mu +\sum_{x\in D_K}{x} ,\frac{1}{1 + b_\mu + ||D_K||}\right),
\end{equation}
where again $D_K$ is the synthetic data, to which we append the current $K$ on \textit{successful} iterations. 

$\blacklozenge$ Similarly, for the size $c$ of the neighbourhood operator $N_c(\boldsymbol{m})$ of SA we assume $c\sim \text{Binomial}(C,p_c)$ with $p_c\sim\text{Beta}(a_c,b_c)$. Then the posterior adaptive distribution for $c$ becomes:
\begin{equation}
p(c|D_c)\overset{d}{=}\text{BetaBin}\left(a_c +\sum_{x\in D_c}{x} ,b_c + C\cdot||D_c|| - \sum_{x\in D_c}{x}\right).
\end{equation}
Here, for \textit{successful} iterations we append $c$ to the synthetic data $D_c$. 

$\blacklozenge$ Finally, for the probabilities of having the covariates included into the model $\Psi_j, j\in \{1,...,p\}$ we consider the Multinomial prior $\Psi_j\sim\text{Multinomial}(p_{j_1},p_{j_2},p_{j_3},p_{j_4})$,  where the four states correspond to $\{\gamma_j=0,\delta_j=0\}$, $\{\gamma_j=1,\delta_j=0\}$, $\{\gamma_j=0,\delta_j=1\}$ and $\{\gamma_j=1,\delta_j=1\}$ accordingly. Here, we set a conjugate Dirichlet prior for the hyperparameters of the Multinomial distribution $\{p_{j_1},p_{j_2},p_{j_3},p_{j_4}\}\sim\text{Dirichlet}(\zeta_1,\zeta_2,\zeta_3,\zeta_4)$ leading to a Dirichlet-Multinomial posterior adaptive distribution: 
\begin{align}
    &p(\Psi_j|D_{\Psi_j})\overset{d}{=}\text{DirMult}(\zeta_1+\sum_{x\in D_{\Psi_j}}{\text{I}(x=\{0,0\})},\nonumber\\ &\zeta_2+\sum_{x\in D_{\Psi_j}}{\text{I}(x=\{0,1\})},\zeta_3+\sum_{x\in D_{\Psi_j}}{\text{I}(x=\{1,0\})},\\ &\zeta_4+\sum_{x\in D_{\Psi_j}}{\text{I}(x=\{1,1\})}).\nonumber
\end{align}
Here, for \textit{successful} iterations we update the synthetic data $D_{\Psi_j}$ with the currently drawn $\Psi_j$. Other tuning parameters of the algorithm including the cooling schedule and the stopping time for adaptive learning are assumed fixed.
%\begin{wrapfigure}{L}{0.469\textwidth}
%\begin{minipage}{0.469\textwidth}
%\end{minipage}
%\end{wrapfigure}
\begin{algorithm}[H]
\tiny %\small
   \caption{\small Adaptive Simulated Annealing Expectation Maximization}
   \label{a1}
\begin{algorithmic}
\STATE \textbf{initialize} $\{\boldsymbol {m^g},\boldsymbol\theta_{m^g}\}$;
   \FOR{$r$ {\bfseries in} $\{1,...,R\}$}
   \STATE \textbf{initialize} $\{\boldsymbol {m^c},\boldsymbol\theta_{m^c}\}$;
   \FOR {$\tau$ {\bfseries in} $\{\tau_{max},...,\tau_{min}\}$}
   \STATE \textbf{sample} $K\sim p(K|D_K)$;
   \FOR{$k$ {\bfseries in} $\{1,...,K\}$}
   \STATE \textbf{sample} $c\sim p(c|D_c)$;
   \STATE \textbf{choose} $J$ from $N_c(\boldsymbol m)$;
   \STATE \textbf{sample} $m^*_j$ from $p(\Psi_j|D_{\Psi})$ for $j\in J$;
   \STATE \textbf{initialize} $\boldsymbol \theta_{m^*}$ for $m^*$;
   \STATE \textbf{sample} $l\sim p(l|D_l)$;
   \STATE \textbf{run} EM for $l$ iterations and \textbf{obtain} $\widehat{\boldsymbol{\theta}}_{m^*}$; 
   \IF{$a_\tau\left(r(\boldsymbol {m^c},\boldsymbol\theta_{m^c}|\boldsymbol Y, \boldsymbol x),r(\boldsymbol m^*,\boldsymbol\theta_{m^*}|\boldsymbol Y, \boldsymbol x)\right)$}
   \STATE \textbf{set} $\{\boldsymbol {m^c},\boldsymbol\theta_{m^c}\}\gets{\{\boldsymbol m^*,\boldsymbol\theta_{m^*}\}}$;
    \IF{$r<R_a$ and $\tau<1$}
    \STATE $D_l\gets{D_l \cup l}$, $D_c\gets{D_c \cup c}$, $D_K\gets D_K \cup K$, $D_\Psi\gets{D_\Psi \cup \Psi}$;
    \ENDIF
   \IF{$r(\boldsymbol {m^c},\boldsymbol\theta_{m^c}|\boldsymbol Y, \boldsymbol x)<r(\boldsymbol m^g,\boldsymbol\theta_{m^g}|\boldsymbol Y, \boldsymbol x)$}
   \STATE \textbf{set} $\{\boldsymbol {m^g},\boldsymbol\theta_{m^g}\}\gets{\{\boldsymbol m^c,\boldsymbol\theta_{m^c}\}}$;
   \IF{$r<R_a$}
    \STATE $D_l\gets{D_l \cup l}$, $D_c\gets{D_c \cup c}$, $D_K\gets D_K \cup K$, $D_\Psi\gets{D_\Psi \cup \Psi}$;
   \ENDIF
   \ENDIF
   \ENDIF
   \ENDFOR
   \ENDFOR
   \ENDFOR
\end{algorithmic}
\end{algorithm}
The ASA-EM Algorithm \ref{a1} converges to a homogeneous Markov chain by design, since the adaptations of the hyperparameters are stopped after $R_a$ epochs. Hence after $R_a$ epochs we end up with a standard simulated annealing algorithm for the model exploration combined with a standard expectation maximization algorithm for parameter estimation. The resulting algorithm is represented by a Markov chain and converges to a stationary distribution at any given temperature $\tau$ independently of the starting solution. 
At the same time, the EM part of the algorithm is greedy and thus it finds only the local optimum, which depends on the starting points. Multiple revisits of the same model $\boldsymbol{m}$ with different initial points for parameters in $\boldsymbol{\Theta}_m$ resolves this issue, if there is only a finite countable number of local extrema of the objective in $\boldsymbol{\Theta}_m$. ASA-EM has a guarantee to find the global optimum asymptotically since it is a positively recurrent Markov chain in $\{\mathcal{M}\bigcup_{\boldsymbol{m}\in\mathcal{M}}\boldsymbol{\Theta}_m\}$. However, in the limiting case, it needs exponential time to do this. Otherwise (within a small finite amount of time) it can be seen as a  metaheuristic optimization strategy for model selection in non-homogeneous hidden Markov models. 

The algorithm can be embarrassingly parallelized in a straight forward fashion either from the beginning (with the same or different tuning parameters for each thread) or after $R_a$ epochs \footnote{\AH{Whilst we do not give any guidance on how many $R_a$ epochs should be considered, we numerically show that even 2-3 $R_a$ epochs are typically enough, provided that at least one additional epoch is run to guarantee ergodicity of the Markov chain}}. \AH{Note that additionally different number of hidden states can be considered in different threads by e.g. sampling from a uniform integer distribution within a reasonable range, allowing to combine variable selection problem with the problem of selecting the optimal number of states.} 

\section{Experiments}
To study the performance of the proposed model and algorithm in the case of limited computational time we will address the data from the first chromosome of Arabidopsis  plant belonging to several predefined groups of genes and the logreturns data from the \emph{S\&P 500} listing. The first example will serve as a pattern recognition study alongside the genome, whilst the second example is used as a prediction driven experiment.

\subsection{Epigenetic data example}

The addressed data set consists of 1502 observations from the first chromosome of Arabidopsis  plant belonging to several predefined groups of genes. The observations are represented by the methylated $Y_i$ versus total $N_i$ number of reads inducing the binomial distribution of the responses. The hidden Markov chain here has just two states (1 - methylated and 0 - non-methylated), which results in the following model specification:
\begin{displaymath}
   Y_i|p_{s_i},s_i,N_i\sim   \text{Binomial}(p_{s_i},N_i),
  \quad p_{{s_i}} = \text{logit}^{-1}\left(\eta_{s_i}\right), s_i\in\{0,1\}. %\label{themodelbin} 
\end{displaymath}
We also have data on various exogenous variables (covariates). Among  these covariates, we address (following \citep{hubin2019}) the factor with 3 levels corresponding to whether the location belongs to CGH, CHH or CHG genetic region, where H is either A, C or T and thus generating two covariates $X_{CG}$ and $X_{CHG}$. The second group of factors indicates whether the distance to the previous  cytosine nucleobase (C) in DNA is 1, 2, 3, 4, 5, from 6 to 20 or greater than 20 inducing six binary covariates $X_{DT1},X_{DT2},X_{DT3},X_{DT4},X_{DT5}$, and $X_{DT6:20}$. We also include such 1D distance as a continuous covariate $X_{DIST}$. The third addressed group of factors corresponds to whether the location belongs to a gene from a particular group of genes of biological interest. These groups are indicated as $M_a$, $M_g$ and $M_d$, yielding two additional covariates $X_{M_a},X_{M_g}$. Additionally, we have a covariate $X_{CODE}$ indicating if the corresponding nucleobase is in the coding region of a gene and a covariate $X_{STRD}$ indicating if the nucleobase is on a "+" or a "-" strand. Finally, we have a continuous covariate $X_{EXPR} \in \mathbb{R^{+}}$ representing expression level for the corresponding gene and interactions between expression levels and gene groups $X_{EXPR,a}, X_{EXPR,g}, X_{EXPR,d} \in \mathbb{R^{+}}$. Thus multiple predictors with respect to a strict choice of the reference model in our example induced $M = 17$ potentially important covariates, yielding $4^{17} > 17\times 10^9$ potential models to consider. %Finally, the correlation structure between the addressed explanatory variables is depicted in Figure~\ref{ex5corr}. We can see multiple significant correlations between the explanatory variables, which might well results in significant sparsification.

\begin{table}
\tiny
  \caption{\small hyperparameters of ASA-EM. Here $\text{U}[a,b]$ is a uniform distribution on the interval from $a$ to $b$.} \label{hyper}
  \begin{tabular}{lllllll}
    \toprule
$\zeta_1$&$\zeta_2$&$\zeta_3$&$\zeta_4$&$a_\lambda$&$b_\lambda$&$a_\mu$\\
0&0&0&1&$100*\text{U}[1,2]$&$\text{U}[1,10]$&$5*\text{U}[3,7]$\\
\midrule
$b_\mu$&$a_s$&$b_s$&$R_a$&$\tau_{min}$&$\tau_{max}$&$k_t$\\
2&5&15&3&$5*\text{U}[10^{-7},10^{-2}]$&$2*\text{U}[10^4,10^9]$&$\text{U}[2,6]$\\
  \bottomrule
\end{tabular}
\end{table}

\begin{table}
\tiny
  \caption{\small The selected model for epigenetics study.} \label{tab:epi}
  \begin{tabular}{cccc}
    \toprule
BIC&AIC&Init. State 1&Init. State 2\\
1481.928&1412.839&1.000&0.000\\
\midrule
$\rightarrow$ State 1&Intercept&$X_{M_a}$&$X_{EXPR}$\\
State 1&0.000&0.000&0.000\\
State 2&-129.264&119.104&0.000\\
 \midrule
$\rightarrow$ State 2&Intercept&$X_{M_a}$&$X_{EXPR}$\\
State 1&0.000&0.000&0.000\\
State 2&5.478&0.288&0.000\\
\midrule
Observations&Intercept&$X_{CHG}$&$X_{CG}$\\
State 1&-5.518&-79.583&-1.770\\
State 2&-2.303&1.714 &2.152\\
  \bottomrule
\end{tabular}
\end{table}

\begin{figure}[h]
    \centering
    \includegraphics[trim={0cm 0cm 1.0cm 1.8cm},clip,width=1\linewidth]{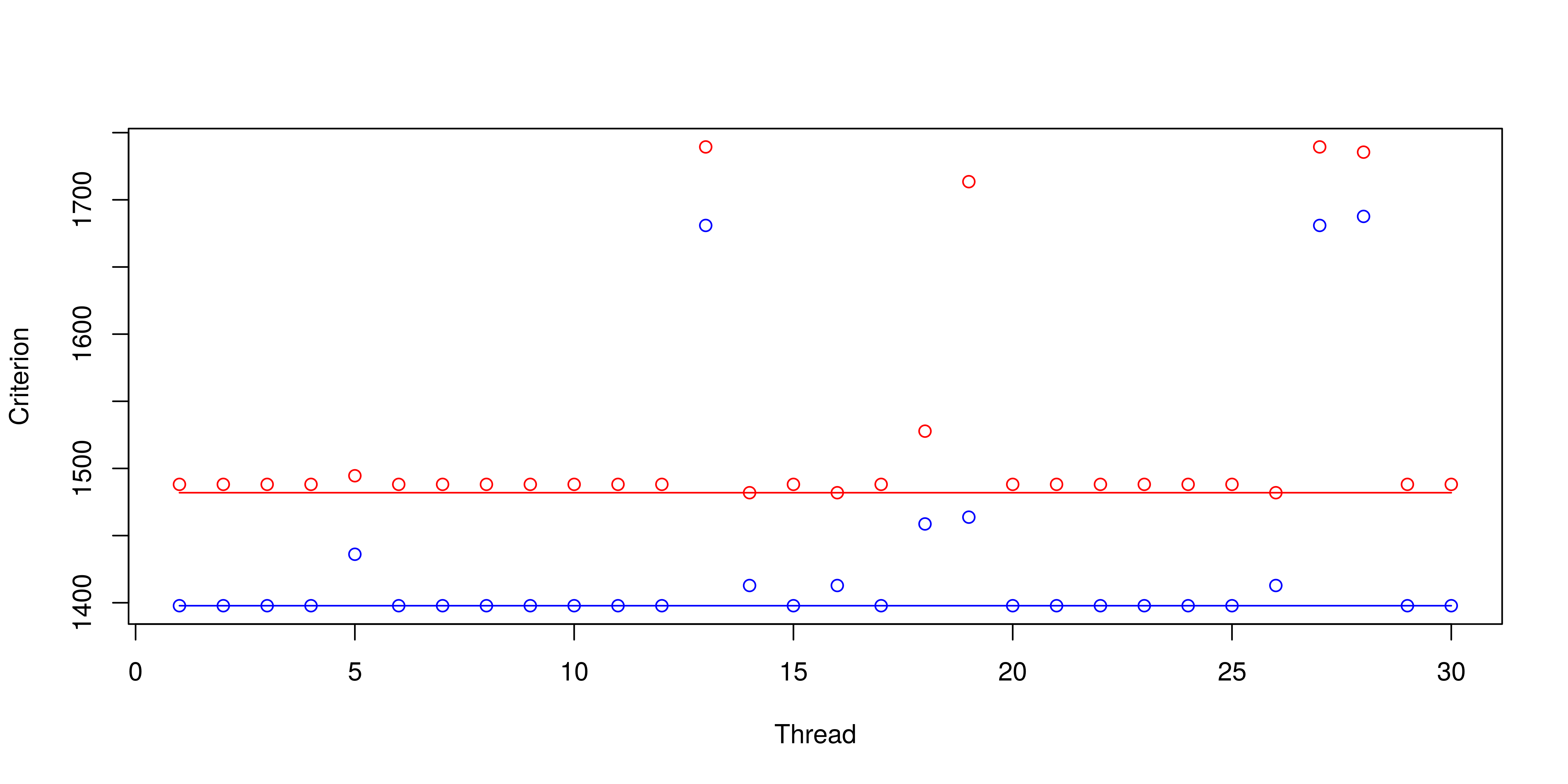}
    \caption{\small Values of BIC (red) and AIC (blue) across 30 parallel threads of ASA-EM algorithm for epigenetics data.}
    \label{modelEPI}
\end{figure}

\begin{figure}[ht]
\centering
\includegraphics[trim={0cm 0cm 1.0cm 1.8cm},clip,width=1\linewidth]{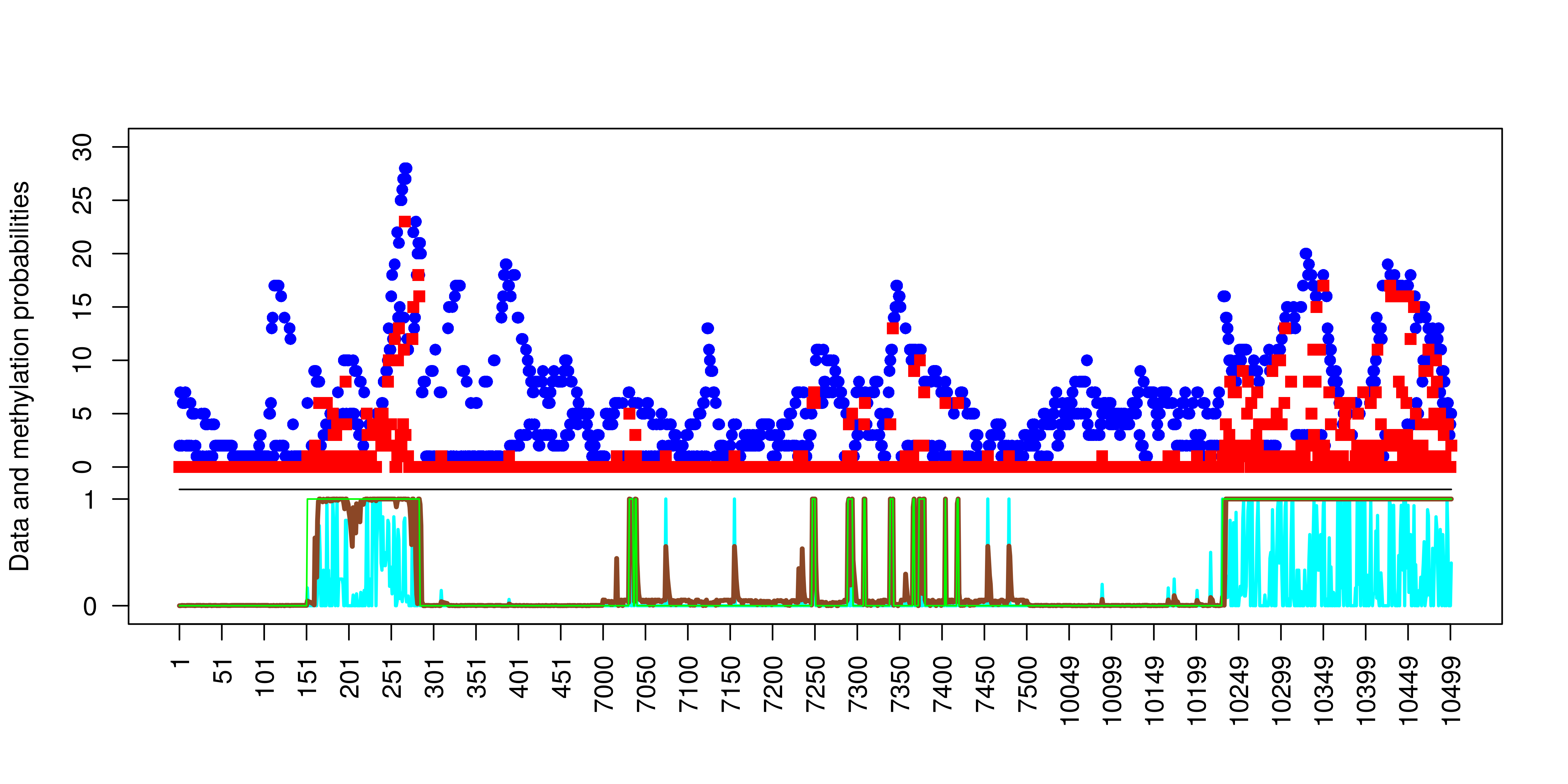}
\caption{\small Epigenetic observations, where blue dots are the total number of reads, red dots - number of methylated reads, the light blue line gives na\"ive probabilities as rates, the green line - most probable latent states of the non-homogeneous hidden Markov model, the brown line - probability of a latent state being 1.}\label{sampleEPI}
\end{figure}

We run 30 parallel threads of ASA-EM algorithm with the choice of hyperparameters shown in Table~\ref{hyper}, whilst $p(s|\theta)$ and sampling in the E-step are the default choices from \citet{depmixs4}. The distribution of the values of model selection criteria (AIC and BIC) across the runs is shown in Figure~\ref{modelEPI} with BIC used as a primary criterion for optimization. Then the best model found in all of the threads in terms of BIC was selected for the given pattern recognition task. According to Figure~\ref{modelEPI}. Exactly this model was found in $10\%$ of the threads, whilst in $87.7\%$ of the threads, the deviation of the BIC value of the found solution is almost neglectable in comparison to the best one found.  The structure of the observations together with the discovered pattern of the most probable sequence of states, representing the methylation status of the observations, found by the suggested ASA-EM procedure, are represented in Figure~\ref{sampleEPI}, whilst the selected model is summarized in Table~\ref{tab:epi}.
% \begin{figure}[h]
%     \centering
%     \includegraphics[width=1\linewidth]{ex5corr.png}
%     \caption{Correlation structure of the covariates in epigenetics study.}
%     \label{ex5corr}
% \end{figure}

According to the BIC criterion, only factors $X_{CHG}$ and $X_{CG}$ are  significant on the level of observations (conditional on the state) of the methylation for the addressed epigenetic region. At the same time, the factor $X_{M_a}$ and the level of expression $X_{EXPR}$ are significant for the transition probabilities between the latent states. The detailed influence on the linear predictors and transition is given in Table~\ref{tab:epi}. Based on the results for the best model we carried out computations of methylation probabilities of the locations along the genome. Furthermore, we compared the results with the na\"ive approach based on computing the proportion of methylated reads, which is currently addressed in the biological literature as a standard way to evaluate methylation probability of a given nucleobase. These results are summarized in Figure~\ref{sampleEPI}. The results show that the na\"ive approach should not be trusted in the presence of spatially correlated data and the corresponding to it probabilities are strongly biased. Previously, a similar analysis of exactly the same data was performed in \citet{hubin2019}, where the inference was based on the GLMM model fitted by the MJMCMC algorithm developed in \citet{hubin2016} also with the aim to capture spatial structure of the methylation probabilities. In \citet{hubin2019} the significant variables of GLMM's linear predictor were  $X_{CHG}, X_{CG}$ and $X_{CODE}$ with less significant $X_{M_a}$ and $X_{M_g}$. Hence variable selection in both of the approaches shows the importance of almost identical sets of covariates, though the models are rather different. This creates insights to be studied further by the biologists. The patterns of the recovered probabilities and most probable states (found by the Viterbi forward algorithm \citep{collins2002discriminative}) in our study and in \citet{hubin2019} are also quite similar and strongly agree for locations 1-7000 and 10250 - 10500. Also, both of the approaches are quite sceptical to the methylation status of locations 7100 -7200, despite some observations with a high proportion of methylated reads in that region. Finally, the approaches disagree in locations 7250 -7400, where the GLMM suggests an almost continuously methylated region, whilst the non-homogeneous HMM models multiple jumps between the methylated and non-methylated states. This region hence might need careful studies by the biologists in future. In would be also of interest to obtain additional covariates such as whether the corresponding nucleobase belongs to a particular part of the non-coding gene region like promoter, intron or transposon, and whether the nucleobase is within a CpG island. 
\subsection{S\&P500 data example} Our second example is focused upon \textit{Amazon} stock (AMZN) prediction with respect to logreturns of other $p = 30$ stocks from the S\&P500 listing, which have the highest correlations to AMZN. The addressed data \footnote{\AH{From \url{www.kaggle.com/camnugent/sandp500}.}} consists of 1258 observations of logreturns for 31 stocks based on the daily close price. The 30 predictors are:
$X_\text{ADBE}$, $X_\text{AOS}$, $X_\text{APH}$, $X_\text{ATVI}$, $X_\text{BLK}$, $X_\text{CA}$, $X_\text{CRM}$, $X_\text{FB}$, $X_\text{FISV}$, $X_\text{GOOGL}$, $X_\text{GOOG}$, $X_\text{ITW}$, $X_\text{MA}$,  $X_\text{MHK}$, $X_\text{MMC}$, $X_\text{MSFT}$, $X_\text{MTD}$, $X_\text{NFLX}$, $X_\text{PCLN}$, $X_\text{PKI}$, $X_\text{ROP}$, $X_\text{SBUX}$, $X_\text{SNPS}$, $X_\text{SPGI}$, $X_\text{SYK}$, $X_\text{TEL}$,  $X_\text{V}$, $X_\text{TMO}$, $X_\text{VRSN}$, where the underscripts are representing official S\&P500 acronyms of the corresponding stocks. In terms of physical time, the observations are ranged from 11.02.2013 to 07.02.2018. The focus of this example is in both inference and predictions, hence we divided the data into a training data set (before 01.01.2017) and a testing data set (after 01.01.2017). Here, we address the Gaussian NHHMM with three classes, which we hypothesize to represent the "buy", "sell", "wait" states of the market:
\begin{displaymath}
   Y_i|\mu_{s_i},\sigma_{s_i}^2,s_i\sim   N(\mu_{s_i},\sigma^2_{s_i}),
  \quad \mu_{{s_i}} = \eta_{s_i}, s_i\in\{0,1,2\}. %\label{themodelbin} 
\end{displaymath}

\begin{figure}[h]
    \centering
    \includegraphics[trim={0cm 0cm 1.0cm 1.8cm},clip,width=1\linewidth]{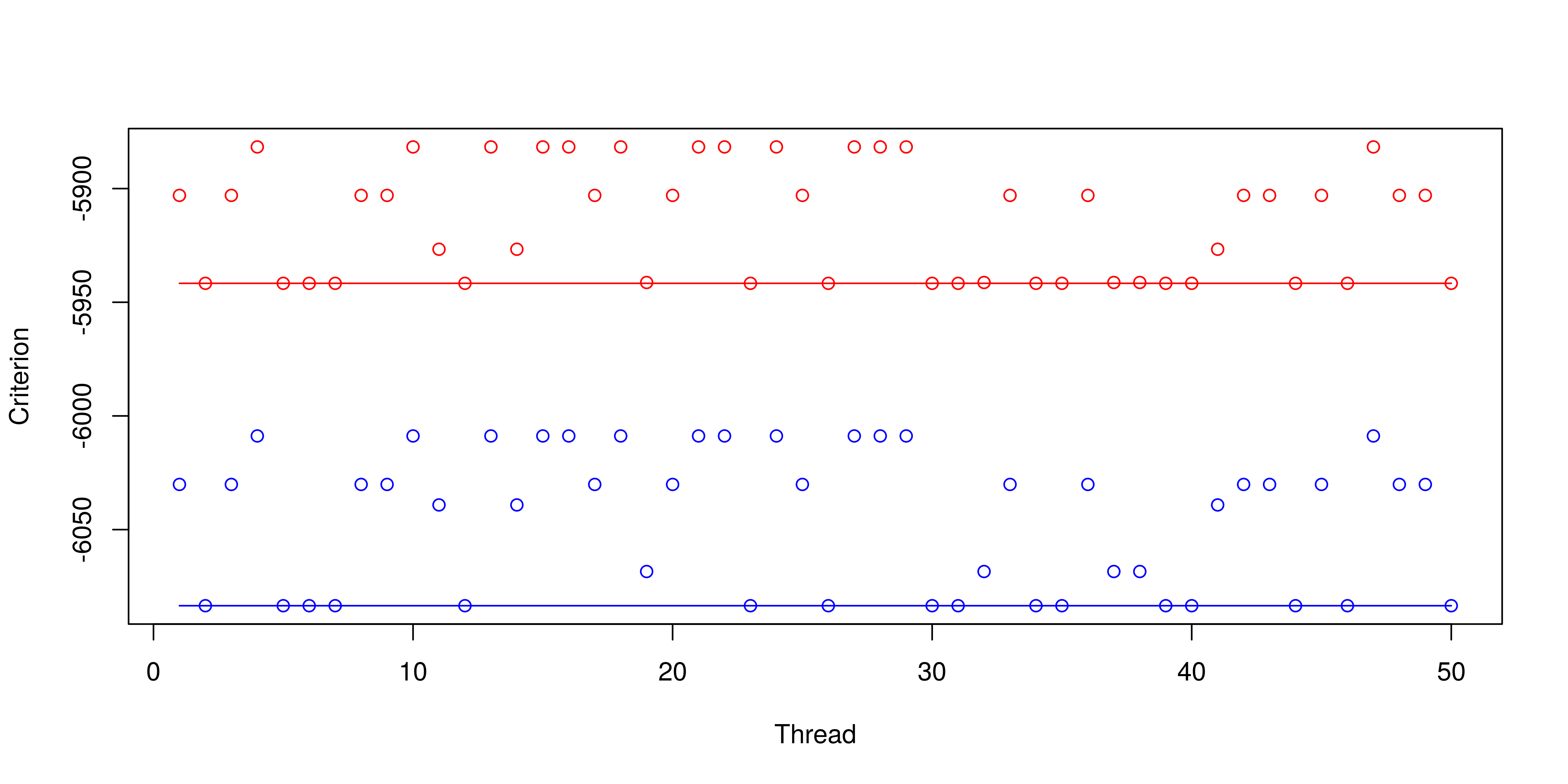}
    \caption{\small Values of BIC (red) and AIC (blue) across 50 parallel threads of ASA-EM algorithm for S\&P500 data.}
    \label{modelSP500}
\end{figure}

\begin{figure}[h]
\centering
\includegraphics[trim={0cm 2.8cm 7.0cm 12.6cm},clip,width=1\linewidth,height =1\linewidth]{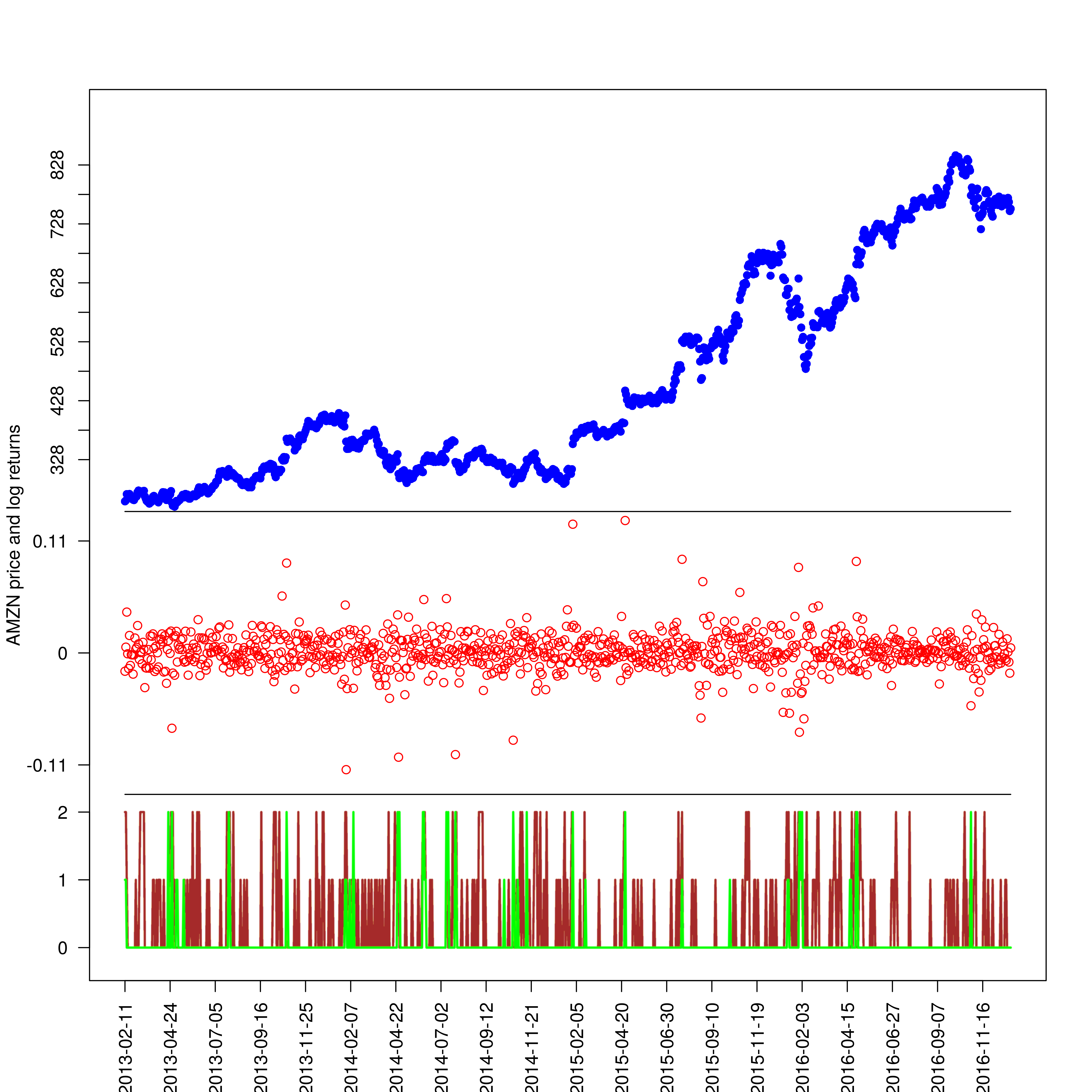}
\caption{\small Amazon prices and logreturns for training data, where blue dots are stock prices, red dots - logreturns, green line - most probable latent states of \AH{the selected by ASA-EM} \AH{NHHMM, brown line - most probable latent states of full NHHMM.}}\label{sampleFIN}
\end{figure}

The primary model selection criterion addressed in this example is AIC due to that we are mainly interested in predictions, whilst BIC is the secondary reported criterion. Other hyperparameters of the algorithm are the same as the ones used in the previous example. Here we have in total $4^{30}>1.15\times 10^{18}$ candidate solutions, hence 50 parallel threads of ASA-EM algorithm were run. The distribution of the best-found solutions across the threads is depicted in Figure~\ref{modelSP500}. We clearly see three major local extrema to which the threads are converging. This indicates that more epochs of each thread would be needed to resolve the issue of convergence to multiple local extrema. On the other hand, it would also require significantly more computational effort. In this sense (and since we are interested in one "best" solution) running more parallel threads (50 instead of 30) is a far more practical and pragmatic solution. The best found model is summarized in Table \ref{tab:fin}.
We use the standard deviations $\sigma_{s_i}, {s_i\in\{0,1,2\}}$ and means $\mu^0_{{s_i}}, {s_i\in\{0,1,2\}}$ at zero values of the predictors for distributions of the observations from the 3 different classes in order to resolve the label switching problem and see if the discovered classes are "buy", "wait", and "sell". These standard deviations and means at zero values of the covariates are $\sigma_0 = 0.008, \mu^0_{{0}} = 0.000$ $\sigma_1 = 0.007,\mu^0_{{1}} = -0.001$ and $\sigma_2 = 0.019, \mu^0_{{2}} = 0.009$. This indicates that $s_i = 0$ might represent the "wait" or neutral state of the market, $s_i=1$ - the "buy" state of the market and $s_i = 2$ - the "sell" state of the market. Let us illustrate the market structure, returns and hidden states graphically. In Figure~\ref{sampleFIN} we clearly see that $s_i=2$ states of \AH{the selected by ASA-EM NHHMM (SNHHMM)} correspond to the pikes of the price and highly volatile logreturns and hence create the best moment for the investor to sell the stock. The $s_i = 1$ states correspond to the moments when the price starts to grow and volatility of the logreturns is low, which can be clearly interpreted as the best moment to buy the stock. The states $s_i = 0$ correspond to the "wait" situation on the market when the investor should hold either the bought asset or the funds. \AH{At the same time, this does not seem to be the case for the full NHHMM (FNHHMM), which has all of the covariates included. FNHHMM has less transparent states in terms of interpretations (hence we will not use states of FNHHMM as investment triggers).} Note that the most probable states for both inference and predictions are recovered by the Viterbi algorithm \citep{collins2002discriminative, depmixs4}.
\setlength{\tabcolsep}{0.1mm}
\begin{table}
\tiny
  \caption{\small The selected model for s\&p500 study of AMZN prediction}
  \label{tab:fin}
  \begin{tabular}{cccccccccccc}
    \toprule
&&&BIC&AIC&Init. State 1&Init. State 2&Init. State 3\\
&&&$-$5534.046&$-$6159.914&0.000&1.000&0.000\\
\midrule
$\rightarrow$ State 1&Inter.&$\tiny X_\text{\tiny BLK}$&$\tiny X_\text{\tiny FISV}$&$\tiny X_\text{\tiny GOOG}$&$\tiny X_\text{\tiny MA}$&$\tiny X_\text{\tiny MHK}$&$\tiny X_\text{\tiny MMC}$&$\tiny X_\text{\tiny 4NFLX}$&$\tiny X_\text{\tiny SBUX}$&$\tiny X_\text{\tiny TEL}$&$\tiny X_\text{\tiny TMO}$\\
State 1&0.00&0.00&0.00&0.00&0.00&0.00&0.00&0.00&0.00&0.00&0.00\\
State 2&$-$5.80&14.66&$-$13.09&$-$197.69&54.81&25.68&$-$9.14&50.23&4.31&$-$73.78&74.03\\
State 3&$-$4.81&6.88&$-$745.00&$-$67.42&$-$32.22&$-$90.12&139.29&$-$5.61&63.93&$-$49.40&49.85\\
\midrule
$\rightarrow$ State 2&Inter.&$\tiny X_\text{\tiny BLK}$&$\tiny X_\text{\tiny FISV}$&$\tiny X_\text{\tiny GOOG}$&$\tiny X_\text{\tiny MA}$&$\tiny X_\text{\tiny MHK}$&$\tiny X_\text{\tiny MMC}$&$\tiny X_\text{\tiny 4NFLX}$&$\tiny X_\text{\tiny SBUX}$&$\tiny X_\text{\tiny TEL}$&$\tiny X_\text{\tiny TMO}$\\
State 1&0.00&0.00&0.00&0.00&0.00&0.00&0.00&0.00&0.00&0.00&0.00\\
State 2&$-$8.75&$-$410.08&$-$838.22&154.20&$-$675.17&449.64&$-$708.36&$-$183.25&639.17&1509.92&548.74\\
State 3&$-$14.84&951.36&866.36&$-$40.28&445.56&210.34&$-$589.45&$-$181.41&$-$128.87&$-$929.12&$-$754.37\\
\midrule
$\rightarrow$ State 3&Inter.&$\tiny X_\text{\tiny BLK}$&$\tiny X_\text{\tiny FISV}$&$\tiny X_\text{\tiny GOOG}$&$\tiny X_\text{\tiny MA}$&$\tiny X_\text{\tiny MHK}$&$\tiny X_\text{\tiny MMC}$&$\tiny X_\text{\tiny 4NFLX}$&$\tiny X_\text{\tiny SBUX}$&$\tiny X_\text{\tiny TEL}$&$\tiny X_\text{\tiny TMO}$\\
State 1&0.00&0.00&0.00&0.00&0.00&0.00&0.00&0.00&0.00&0.00&0.00\\
State 2&$-$0.29&494.63&$-$800.14&1449.02&758.39&387.67&$-$652.24&$-$366.78&266.96&$-$1132.37&$-$372.49\\
State 3&$-$12.66&192.90&223.25&1463.84&323.61&930.92&$-$1671.98&$-$132.73&626.89&$-$277.53&$-$383.94\\
\midrule
Observ.&Inter.&$\tiny X_\text{\tiny ADBE}$&$\tiny X_\text{\tiny APH}$&$\tiny X_\text{\tiny ATVI}$&$\tiny X_\text{\tiny BLK}$&$\tiny X_\text{\tiny CA}$&$\tiny X_\text{\tiny FB}$&$\tiny X_\text{\tiny GOOGL}$&$\tiny X_\text{\tiny ITW}$&$\tiny X_\text{\tiny MA}$&$-$\\
State1&0.00&0.10&0.03&0.05&0.00&$-$0.02&0.11&0.42&$-$0.07&0.09&$-$\\
State 2&0.00&$-$0.11&$-$0.72&0.08&0.17&$-$0.64&$-$0.04&$-$0.13&$-$0.52&0.89&$-$\\
State3&0.01&$-$1.32&$-$0.07&0.30&$-$1.56&0.38&$-$1.03&3.01&1.64&$-$0.85&$-$\\
Observ.&SD&$\tiny X_\text{\tiny MMC}$&$\tiny X_\text{\tiny NFLX}$&$\tiny X_\text{\tiny PCLN}$&$\tiny X_\text{\tiny SBUX}$&$\tiny X_\text{\tiny SNPS}$&$\tiny X_\text{\tiny SYK}$&$\tiny X_\text{\tiny TEL}$&$\tiny X_\text{\tiny VRSN}$&$\tiny X_\text{\tiny V}$&$-$\\
State 1&0.01&0.03&0.05&0.08&0.15&0.02&0.03&$-$0.02&0.11&$-$0.03&$-$\\
State 2&0.01&0.12&$-$0.23&0.37&$-$0.06&0.52&0.32&0.25&0.36&1.16&$-$\\
State 3&0.02&3.58&0.18&$-$0.02&$-$0.15&0.93&0.50&$-$1.77&$-$0.65&$-$0.91&$-$\\
  \bottomrule
\end{tabular}
\end{table}
Consider that the investor during the training period keeps to the following strategy: She invests all her money into AMZN stock when the latent variable \AH{of SNHHMM} takes the value of 1 and sells all her stocks when the latent variable takes the value of 2. Assume the investor had 1000\$ on 11.02.2013. Then on 01.01.2017 she would have 1915.137\$ in case she used the strategy purely based on SNHHMM, which shows that the latent states indeed correspond to  the "buy", "sell", "wait" states. %2375.171

The goal of this example was not only inference but also predictions, which means we have to check both how the predicted values of AMZN stock price behave and what would have happened should the investor had used the suggested investment strategy on the test data. In terms of the second goal, if the investor kept to the strategy described in the previous paragraph during the test period (from 01.01.2017 to 07.02.2018), she would end up with 1198.636\$ at the end of the test period, provided that she started with 1000\$ on 01.01.2017. This clearly shows that the strategy works in the very long test period, even though both the model and its parameters were fixed on 01.01.2017 and were never updated during test time.

In terms of the other goal, we predict daily stock prices for the whole test period based on the \AH{selected by ASA-EM optimal model (SNHHMM) and full NHHMM (FNHHMM)}. Here, we also address several benchmark approaches, namely LASSO \citep{tibshirani1996regression} and RIDGE regressions \citep{hoerl1970ridge} combined with AIC used for the hyperparameter selection and SARIMAX model with automatic selection of hyperparameters based on AICc criterion \citep{hyndman2007automatic}. \AH{Finally, as two industrial state-of-the-art models we addressed Bayesian structural time series model (CAUSAL) developed by Google \citep{brodersen2015inferring} and PROPHET model developed by Facebook \citep{taylor2018forecasting}\footnote{No explanatory variables are addressed by PROPHET due to its model assumptions.}.} The benchmark approaches were trained and tested using the same splits of the data. For all of the models, the predicted prices of AMZN shares were obtained using the following transformation of logreturns $\widehat{p}_i=p_0\exp(\sum_{j=1}^i\widehat{Y}_j),i\in \{1,...,n_\text{test}\}$, where $p_0$ is the actual stock price on the last training day. Then test RMSE was evaluated as:
\begin{displaymath}
\text{RMSE} = \sqrt{\tfrac{1}{n_{\text{test}}}\sum_{j=1}^{n_{\text{test}}}\left(p_j-\widehat{p}_j\right)^2}.
\end{displaymath}
\AH{Historically stock price prediction algorithms are compared in terms of RMSE. However, relying only on RMSE might be not enough to decide which of them yields the best predictive performance. This is due to the fact that the variance of squared errors of predictions is completely ignored in such a case. Hence rigorous statistical testing should be applied. Simple t-test or z-test in this context can not be trusted, since forecast errors may be serially correlated, and a robust Diebold-Mariano (DM) test \citep{diebold2015comparing}, designed to compare forecasts, should be applied. We hence are additionally reporting p-values (PVDM) of one-sided DM-test \citep{diebold2002comparing} of that our method is superior to the corresponding competitors.} We also tested the absence of auto-correlations of the residuals of the predicted prices using the following cointegration test statistics:
\begin{displaymath}
\text{COINT} = \max\{|\hat{\rho} + 1.96\hat{\sigma}_\rho|,|\hat{\rho} - 1.96\hat{\sigma}_\rho|\}. 
\end{displaymath}
Here  $\hat{\rho}$ is estimated from the following equation $\varepsilon_t = \rho\varepsilon_{t-1} + \eta_t,$
and $\varepsilon_t$ is the noise equal to $\widehat{p}_i - p_{i}$. We want $\text{COINT}$ as small as possible and accept the series to be cointegrated if it is below 1. 
\setlength{\tabcolsep}{0.9mm}
\begin{table}[]
    \centering
    \tiny
    \caption{\small Test statistics results. For DM-test 0.1 level of significance corresponds to $^*$, 0.05 - to  $^{**}$, 0.01 - to  $^{***}$.}
    \begin{tabular}{l|l|lllllll}
    \toprule
    Criterion&Data&NHHMM&\AH{FNHHMM}&LASSO&RIDGE&SARIMAX&\AH{CAUSAL}&\AH{PROPHET}\\
    \midrule
    RMSE&AMZN&\textbf{78.614}&153.391&92.266&89.015&87.439&159.687&181.475\\
    \AH{PVDM}&\AH{AMZN}&\AH{-}&\AH{0.0000$^{***}$}&\AH{0.0109$^{**}$}&\AH{0.0597$^{*}$}&\AH{0.0496$^{**}$}&\AH{0.0000$^{***}$}&\AH{0.0000$^{***}$}\\
    COINT&AMZN&\textbf{0.9999}&1.0043&1.0035&1.0036&1.0048&1.0026&1.0039\\
    MONEY&AMZN&\textbf{1198.63}&-&-&-&-&-&-\\
    RMSE&AAPL&22.795&36.625&19.725&18.738&18.304&\textbf{17.422}&23.601\\
    \AH{PVDM}&\AH{AMZN}&\AH{-}&\AH{0.0000$^{***}$}&\AH{1.0000}&\AH{0.9998}&\AH{1.0000}&\AH{1.0000}&\AH{0.1738}\\
    COINT&AAPL&\textbf{1.0019}&1.0055&1.0055&1.0054&1.0053&1.0050&1.0050\\
    MONEY&AAPL&\textbf{1456.46}&-&-&-&-&-&-\\
    RMSE&AAPL (HALF)&\textbf{5.4772}&33.468&14.333&13.474&13.195&13.124& 20.211\\
    \AH{PVDM}&\AH{AAPL (HALF)}&\AH{-}&\AH{0.0000$^{***}$}&\AH{0.0000$^{***}$}&\AH{0.0000$^{***}$}&\AH{0.0000$^{***}$}&\AH{0.0000$^{***}$}&\AH{0.0000$^{***}$}\\
    COINT&AAPL (HALF)&\textbf{0.9766}&1.0063&1.0105&1.0105&1.0104&1.0102&1.0094\\
    MONEY&AAPL (HALF)&\textbf{1185.59}&-&-&-&-&-&-\\
   % MSE&AAPL&22.795&19.725&18.738&18.304\\
%    COINT&AAPL&1.002&1.005&1.005&1.005 \\
 %   MONEY&AAPL&1205.435&-&-&-\\
    \bottomrule
    \end{tabular}
    \label{resFIN}
\end{table}
The results for the set of compared approaches are summarized in Table~\ref{resFIN}. They clearly show that our method \AH{(SNHHMM)} yields the smallest test RMSE error \AH{and its advantage of predictive performance is statistically significant according to DM-test on $\alpha=0.01$ level for FNHHMM, CAUSAL and PROPHET, on $\alpha=0.05$ level for LASSO and SARIMAX, and on $\alpha = 0.1$ level for RIDGE}. Moreover, SNHHMM is the only method with stationary and cointegrated residuals for AMZN data.
We repeated the experiment for Apple data (AAPL) as observations with 30 explanatory stocks (with the highest correlations to AAPL) and also report the results in Table~\ref{resFIN}. For AAPL, RMSE of SNHHMM was slightly worse than for most of the competitors \AH{except for FNHHMM and  PROPHET} \AH{(also, no significant evidence of that our method is better in accordance with PVDM, except for FNHHMM on $0.01$ level of confidence)}, but cointegration results were still the best. Graphically the results for AAPL are summarized in Figure~\ref{testFIN1}. There one can see that the predictions for SNHHMM worsen in the second part of the test period, where some sort of a causal impact could have happened. This means that the model should be calibrated with some frequency. In particular, the results for all of the methods could be improved should we retrain the models after each test day by adding the corresponding logreturns to improve the following day's predictions, however, we leave this option outside of the scope of the paper. Note, however, that if one only used the first half of the test period for AAPL, both the RMSE \AH{(with a significant difference of predictive performance on $\alpha = 0.001$ level according to DM-test)}, and COINT of SNHHMM would be by far the best on the given set of baselines (see Table~\ref{resFIN}). Finally, using 1000\$ as the starting capital, the investor would gain 1456.457\$ for AAPL using our strategy based on the hidden states during the whole test period and 1185.59\$ during the first half of the period. 
 \begin{figure}[H]
\centering
\includegraphics[trim={0cm 2.8cm 7.0cm 12.6cm},clip,width=1\linewidth,height =1\linewidth]{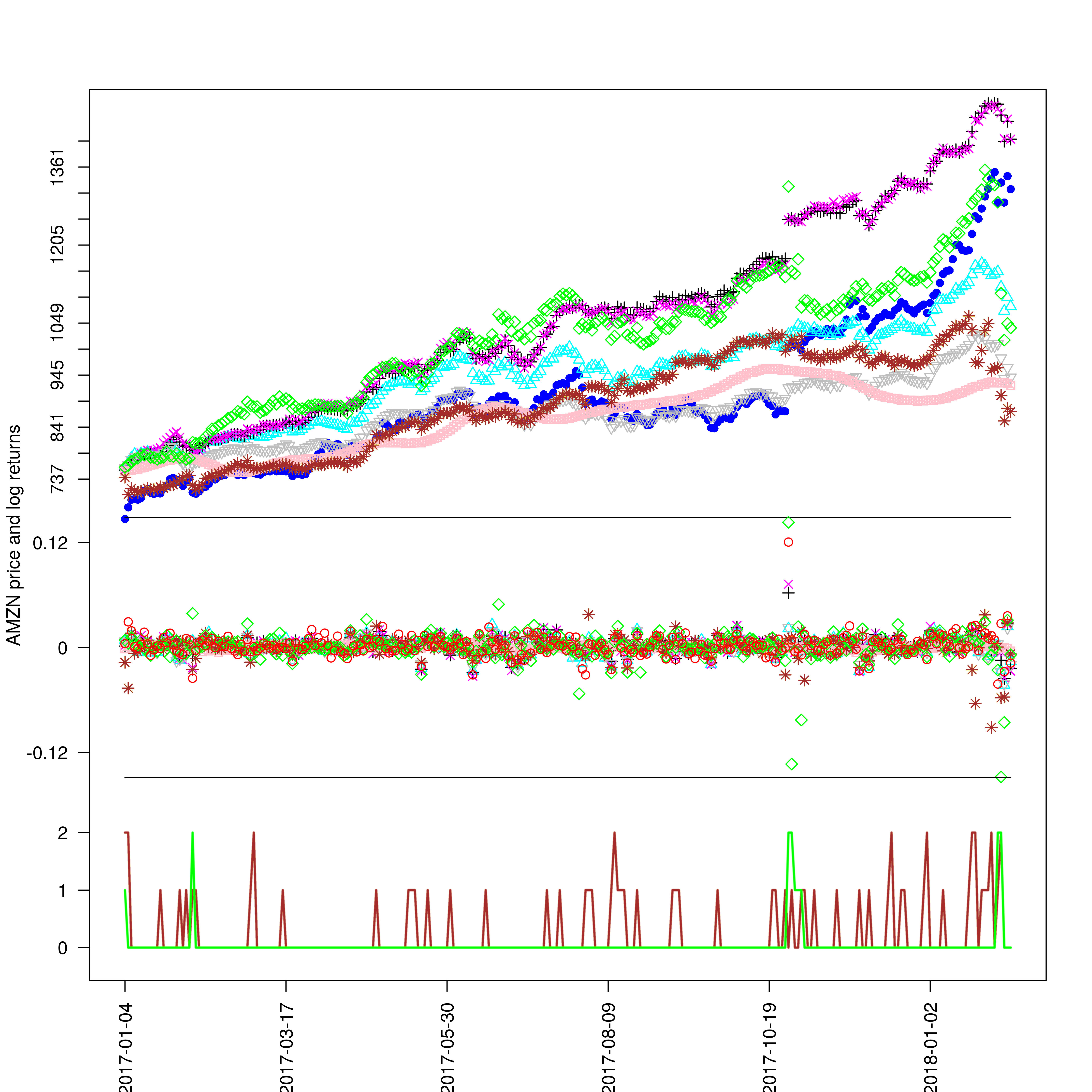}
\caption{\small Amazon prices, log-returns and most probable latent states as in caption to Figure~\ref{sampleFIN}. Green signs correspond to \AH{the selected by ASA-EM} NHHHM predictions, \AH{brown - to full NHHMM,} black - to LASSO, purple - to RIDGE, light blue - to SARIMAX, \AH{grey - to CAUSAL, pink - to PROPHET.}}\label{testFIN}
\end{figure}
\begin{figure}[h]
\centering
\includegraphics[trim={0cm 2.8cm 7.0cm 12.6cm},clip,width=1\linewidth,height =1\linewidth]{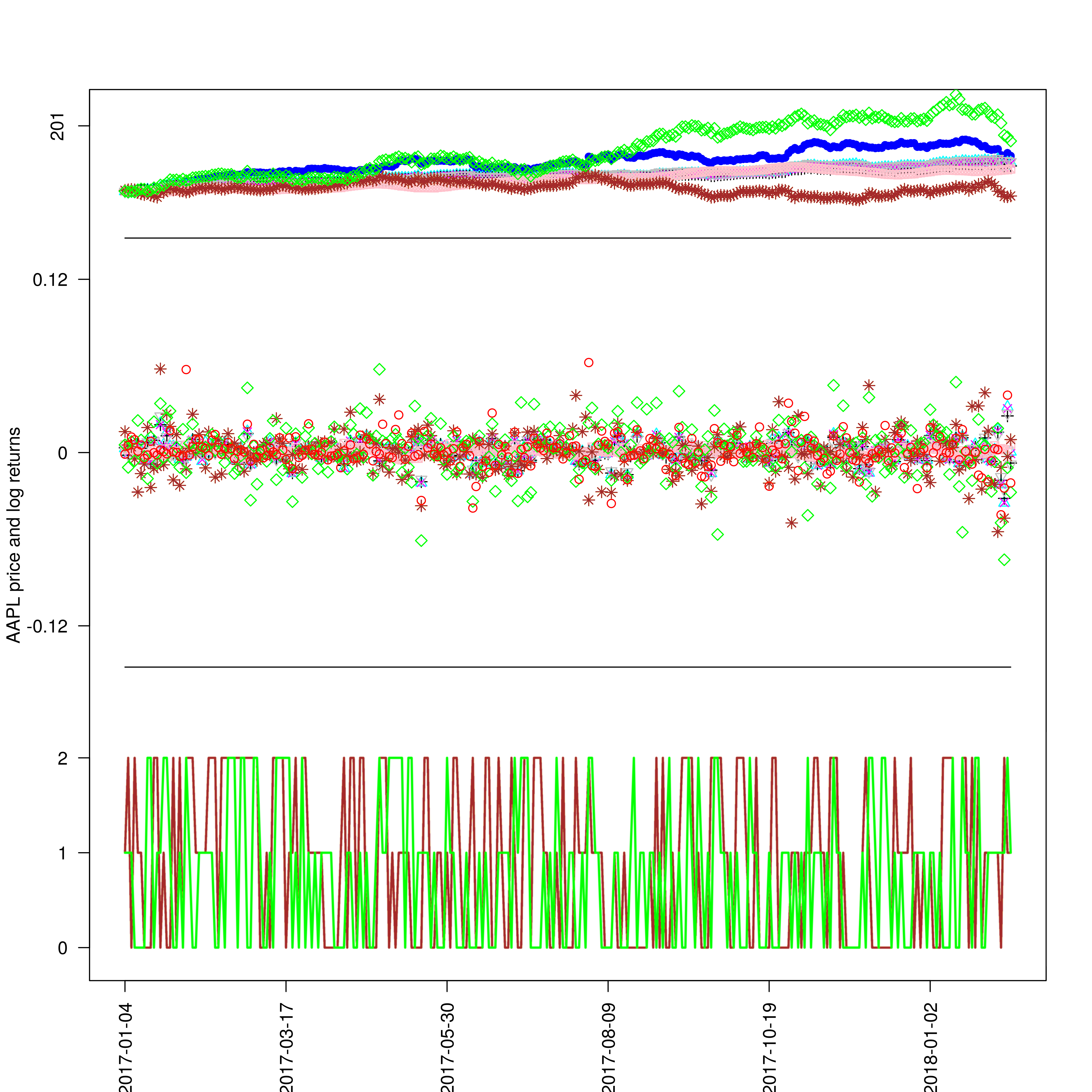}
\caption{\small Apple results. See details in the caption to Figure~\ref{testFIN}}\label{testFIN1}
\end{figure}
\section{Discussion}
In this article, we have suggested a novel adaptive simulated annealing expectation maximization algorithm for simultaneous exploration of model and parameter space of non-homogeneous hidden Markov models. Also, parallelization strategies were suggested for ASA-EM.
The algorithm has shown to efficiently converge to good solutions within finite time. The solutions both yield high sparsification rates and interpretable results in terms of the meaning of the latent states and inference on the observations. In two applications we found the latent states to be associated with important biological and financial phenomena, namely methylation status of the nucleobases along the genome and extremal/non-extremal points of the stock price, which can be interpreted as "buy"-"wait"-"sell" states. Finally, the resulting model yields good predictive properties and often outperforms the baseline approaches. In future, it would be of interest to allow for sub-sampling from the data in the EM part of the algorithm to \AH{deal with large high dimensional} data samples. Also, allowing complex non-linear relations in the model like those described in \citep{hubin2018a} and \citep{hubin2018b} could be of interest.
\footnotesize
\begin{acks}
I would like to acknowledge Professor G. Storvik (UiO) for the discussion of the paper in the first year of my PhD studies. The work resulted in
a talk at Kl{\ae}kken workshop (\url{bit.ly/2MaN3l7}) back in 2015, but then was put aside for a while. The project got a second birth in 2019 inspired by the work on Exabel (\url{exabel.com}) project at NR in collaboration with Dr K. Aas (NR) and Dr P.C. Moan (Exabel), which also led to the experimental design of the S\&P500 example. The epigenetic example was inspired by the discussions with Professors P. Grini, O.C. Lingj{\ae}rde, G. Storvik, K. S. Jakobsen  and Dr M. Butenko (all from UiO). Finally, I would like to thank Dr. P. Lison (NR) for thoroughly proofreading the paper. 

\end{acks}
\small
\bibliographystyle{ACM-Reference-Format}
\bibliography{sample-base}

\end{document}